\documentclass{esannV2}
\usepackage{graphicx}
\usepackage{amssymb}
\usepackage{amsmath}
\usepackage{booktabs}       
\usepackage{mathtools}
\usepackage[group-separator={,},separate-uncertainty=true]{siunitx}
\usepackage{bm} 
\usepackage{xcolor}
\usepackage[final]{microtype}

\newcommand{\rv}[1]{\bm{{#1}}}
\newcommand{\E}{\mathbb{E}}
\DeclareMathOperator{\Dir}{Dir}

\DeclareMathOperator{\BB}{BB}

\newcommand{\Do}{\mathrm{do}}
\newcommand{\e}[1]{^{(#1)}} 

\title{Identifying counterfactual probabilities using bivariate distributions and uplift modeling\thanks{Submitted to the 34th European Symposium on Artificial Neural Networks, Computational Intelligence and Machine Learning.}}

\author{Théo Verhelst$^1$\thanks{Research carried out during the author's prior affiliation with the Machine Learning Group.} \;and Gianluca Bontempi$^2$
\vspace{.3cm}\\
1- Advanced Concepts Team - European Space Agency \\
Noordwijk, Netherlands
\vspace{.1cm}\\
2- Machine Learning Group - Université Libre de Bruxelles \\
Brussels, Belgium\\
}

\begin{document}

\maketitle

\begin{abstract}
Uplift modeling estimates the causal effect of an intervention as the difference between potential outcomes under treatment and control, whereas counterfactual identification aims to recover the joint distribution of these potential outcomes (e.g., “Would this customer still have churned had we given them a marketing offer?”). This joint counterfactual distribution provides richer information than the uplift but is harder to estimate. However, the two approaches are synergistic: uplift models can be leveraged for counterfactual estimation. We propose a counterfactual estimator that fits a bivariate beta distribution to predicted uplift scores, yielding posterior distributions over counterfactual outcomes. Our approach requires no causal assumptions beyond those of uplift modeling. Simulations show the efficacy of the approach, which can be applied, for example, to the problem of customer churn in telecom, where it reveals insights unavailable to standard ML or uplift models alone.
\end{abstract}

\section{Introduction}
Uplift modeling estimates individual-level causal effects (e.g., the impact of showing an ad to a specific user) by training predictors of the potential-outcome probabilities under alternative actions. Such models are typically trained on data from randomized campaigns, where the action assignment is randomized and outcomes are recorded. While running large-scale randomized studies can be costly in domains like medicine, they are comparatively inexpensive in digital settings (e.g., online marketing), where individual-level descriptors and outcomes are readily available. An uplift model trained on these data can then guide future campaigns by targeting users more effectively.

Counterfactual statements concern outcomes under hypothetical alternatives to the realized action. For example: ``This customer churned because we did not make a marketing offer; had we made one, they would have stayed.'' Counterfactuals are used across domains, from algorithmic recourse~\cite{karimi2021algorithmic} to online advertising and customer relationship management~\cite{li2019unit}. Although uplift and counterfactuals are related, they are formally distinct: the counterfactual distribution characterizes probabilities of all realized/hypothetical outcome combinations, whereas uplift captures the change in outcome probability due to the action. The counterfactual distribution is strictly more informative yet typically harder to estimate. As noted by~\cite{li2019unit}, the two notions coincide under monotonicity (absence of negative effects).

Without structural assumptions, counterfactual probabilities are not identifiable from data alone; however, one can derive bounds, a task called \emph{partial counterfactual identification}~\cite{tian2000probabilities}. One line of literature derives such bounds under causal assumptions by combining experimental and observational data~\cite{tian2000probabilities, mueller2021causes, zhang2022partial}. In a complementary direction,~\cite{verhelst2023partial} use uplift scores learned from randomized campaigns to derive bounds on counterfactual probabilities and propose two point estimators: one based on an independence assumption between the potential outcomes, and one given by the midpoint of the bounds. We go further and adopt a Bayesian approach, fitting a parametric joint distribution over the uplift scores; posterior inference in this model yields a posterior distribution over counterfactual probabilities, from which point estimates can be derived. Compared to the baseline estimators of~\cite{verhelst2023partial}, our approach accommodates a broader class of joint distributions (and thus richer dependence structures) over the potential outcomes, which in turn yields more precise estimates of counterfactual probabilities. This uplift-based approach is supported by a mature literature on uplift modeling~\cite{gutierrez2017causal} and is particularly effective when combined with modern machine learning models. Furthermore, it is attractive in applications where graph-based assumptions are hard to justify but large-scale randomized campaigns are feasible.

The main contributions of this paper are as follows: (a) a posterior distribution on the counterfactual probabilities based on a bivariate beta distribution; (b) three variations of this approach that make less restrictive assumptions at the cost of higher computational complexity; (c) an assessment with two different simulations of the proposed counterfactual estimators.

\section{The Bayesian counterfactual approach}
\label{sec:summary}

We denote random variables in boldface (e.g., $\rv x$, $\rv y$, $\rv t$) and their realizations by $x$, $y$, and $t$. In uplift modeling, $\rv y\in\{0,1\}$ is the binary outcome, $\rv t\in\{0,1\}$ is the binary treatment/action, and $\rv x\in\mathcal X\subseteq\mathbb R^n$ is the feature vector. We denote their domains by $\mathcal Y=\{0,1\}$, $\mathcal T=\{0,1\}$, and $\mathcal X\subseteq\mathbb R^n$. We use Pearl's do-notation for causal interventions~\cite{pearl2009causality}: the potential outcome of $\rv y$ under the intervention $\Do(\rv t=t)$ is denoted $\rv y_t$. An uplift model estimates the individual-level effect
\begin{align}
    \label{eq:uplift}
    U(x)=P(\rv y_0=1\mid\rv x=x)-P(\rv y_1=1\mid\rv x=x).
\end{align}

A \emph{counterfactual expression} is any logical expression involving multiple potential outcomes~\cite{bareinboim2022pearl}. In our motivating telecom-churn application, we encode churn as $\rv y=1$ (and no churn as $\rv y=0$), a marketing incentive as $\rv t=1$ (and no incentive as $\rv t=0$), and customer features as $\rv x=x$. Four categories of customers can be delineated depending on the values of their potential outcomes $\rv y_0$ and $\rv y_1$, as shown in Table~\ref{tab:customer_categories}. We focus on the following counterfactual probabilities:
\begin{align}
    p_{ij}&=P(\rv y_0=i,\rv y_1=j) \quad\forall i,j\in\{0,1\}, \label{eq:p} \\ 
    p_{ij}(x)&=P(\rv y_0=i,\rv y_1=j\mid\rv x=x). \label{eq:p_x}
\end{align}
The \emph{population-level} counterfactual $p_{ij}$ is the probability that a randomly drawn individual would have potential outcomes $(\rv y_0,\rv y_1)=(i,j)$, that is, $\rv y=i$ under $\rv t=0$ and $\rv y=j$ under $\rv t=1$. Likewise, the \emph{individual-level} counterfactual $p_{ij}(x)$ is the same probability conditional on $\rv x=x$. 

\begin{table}
    \centering
    \caption{The four categories of customers in terms of potential outcomes.}
    \label{tab:customer_categories}
    \begin{tabular}{c|cc}
        & $\rv y_0=0$ & $\rv y_0=1$ \\[1pt]
        \hline
        \rule{0pt}{1\normalbaselineskip}
        $\rv y_1=0$ & Sure thing & Persuadable\\
        $\rv y_1=1$ & Do-not-disturb & Lost cause
    \end{tabular}
\end{table}

To estimate the probabilities $p_{ij}(x)$ and $p_{ij}$, we assume access to an uplift model trained on data coming from a randomized campaign. The uplift model provides estimates of the scores 
\begin{equation}
    \label{eq:z_0_x}
    z_0(x)=P(\rv y_0=1\mid\rv x=x)\quad\text{and}\quad z_1(x)=P(\rv y_1=1\mid\rv x=x)
\end{equation}
for any $x\in\mathcal X$. We further assume that the scores $z_0(\rv x),z_1(\rv x)$ (note that these are random variables, due to the random nature of the features $\rv x$) follow a specific joint distribution, whose structure allows us to infer the counterfactual probabilities. Our approach relies on the identities
\begin{equation}
    \label{eq:id_z_p}
    z_0(\rv x)=p_{10}(\rv x)+p_{11}(\rv x)\quad\text{and}\quad z_1(\rv x)=p_{01}(\rv x)+p_{11}(\rv x),
\end{equation}
which follow directly from the definition of $z_i(x)$ and $p_{ij}(x)$ (Eqs.~ \ref{eq:p_x} and \ref{eq:z_0_x}). We fit the bivariate beta distribution developed by~\cite{olkin2015constructions} on the joint distribution of scores $z_0(\rv x),z_1(\rv x)$ predicted by the uplift model. This distribution, noted $(\rv\zeta_0,\rv\zeta_1)\sim\BB(m)$ for a vector of positive parameters $m=[m_{00},m_{10},m_{01},m_{11}]$, is a continuous bivariate distribution with beta marginals. Under the hood, it models first the four counterfactual probabilities $p_{ij}(x)$ by sampling them from a Dirichlet distribution $[\rv\pi_{ij}]\sim\Dir(m)$, and then returns the scores $\rv\zeta_0,\rv\zeta_1$ according to the identity \eqref{eq:id_z_p} from $[\rv\pi_{ij}]$.

Our approach consists of two phases:
\begin{enumerate}
    \item
    \textbf{Learning phase:} we train an uplift model on a campaign dataset $D=\left\{\left(x\e i,y\e i,t\e i\right)\right\}_{i=1}^N$ i.i.d. from $(\rv x,\rv y,\rv t)$, resulting in estimators $\widehat z_t\left(x\e i\right)$ of the probability score $z_t\left(x\e i\right)$ (with $t=0,1$). Applying these two estimators on $D$, we obtain $S=\left\{\left(\widehat z_0\left(x\e i\right),\widehat z_1\left(x\e i\right)\right)\right\}_{i=1}^N$. Class imbalance is mitigated with the EasyEnsemble strategy~\cite{liu2009exploratory}, and the scores are re-calibrated using the formula proposed in~\cite{dalpozzolo2015calibrating}. Then, we fit a bivariate beta distribution $\BB(m)$ on the set of scores $F$ using the method of moments.

    \item
    \textbf{Inference phase:} Population-level counterfactuals are estimated using the fitted parameters of the bivariate beta distribution:
    \begin{equation}
        p_{ij}=\E[p_{ij}(\rv x)]\simeq\E[\rv\pi_{ij}]=\frac{m_{ij}}M
    \end{equation}
    where $m_{ij}$ is a parameter of the bivariate beta distribution, and $M=m_{00}+m_{10}+m_{01}+m_{11}$.
    
    Individual-level counterfactual probabilities for a new sample $x'$ are estimated by computing the posterior probability distribution of $\rv\pi_{00},\dots,\rv\pi_{11}$ given the observed scores $\rv\zeta_0=\widehat z_0(x')$ and $\rv\zeta_1=\widehat z_1(x')$. Our proposed point estimator of $p_{00}(x'),\dots,p_{11}(x')$ is the expected value of this posterior distribution. It is estimated by integrating numerically a formula derived from the PDF of the bivariate beta distribution given in~\cite{olkin2015constructions}. For more details on this procedure, we refer the reader to~\cite[Sec.~6.4.3]{verhelst2024causal}.
\end{enumerate}

We also propose three variations of this approach, with varying degrees of additional complexity.

\textbf{Generalized Dirichlet distribution}\quad
The bivariate beta distribution ensures that the scores have marginal beta distributions, and it puts constraints on the possible covariance structures between the counterfactuals $\rv\pi_{ij}$. We alleviate these constraints by replacing the underlying Dirichlet distribution on $\rv\pi_{ij}$ by the \emph{generalized Dirichlet distribution}~\cite{connor1969concepts}, which is a generalization of the Dirichlet distribution with additional parameters. The resulting distribution on the scores $\rv\zeta_t$ is named the \emph{generalized bivariate beta distribution} (GBB for short).\footnote{This is a slight abuse of terminology, since the marginals do not necessarily have beta distributions.}

\textbf{Noisy predictions}\quad
Because the training set is finite, the estimated uplift scores $\widehat z_0(x)$ and $\widehat z_1(x)$ are noisy estimates of $z_0(x)$ and $z_1(x)$. Yet, in the inference phase, we initially treat $z_0(x)$ and $z_1(x)$ as if they were known exactly. To relax this assumption, we introduce a hierarchical Bayesian extension of the bivariate beta distribution, in which beta noise is added on top of the latent scores $\rv\zeta_0,\rv\zeta_1$. The noisy scores are centered on the latent values, with a tunable spread parameter that can be adjusted to reflect the size of the training set. We refer to the resulting distribution as the \emph{noisy bivariate beta distribution}.

\textbf{Generalized Dirichlet distribution with noisy predictions} \quad
Observe that the two variations presented above pertain to different aspects of the bivariate beta distribution, and therefore can be used in tandem, resulting in the \emph{noisy generalized bivariate beta distribution}.

\section{Performance assessment}

In this section, we assess our proposed estimator with simulated data. Using simulated data allows us to compare the estimators with the ground truth, which is not feasible with real data. We consider two data-generating processes: one based on a Gaussian distribution and one based on a bivariate beta distribution. In the Gaussian-based simulation, the features $\rv x$ are sampled from a multivariate standard normal distribution, and the potential outcomes $\rv y_0,\rv y_1$ are generated via a linear combination of $\rv x$ with fixed, randomly chosen weights. In the second simulation, we first fix the parameters of a bivariate beta distribution, then sample pairs of scores from this distribution. We add noise using a beta distribution, as in the noisy bivariate beta model. More details of the experimental protocol are provided in~\cite[Sec.~6.5]{verhelst2024causal}. The bivariate beta simulation is more flexible than the Gaussian simulation and satisfies the modeling assumptions of our counterfactual estimators; the Gaussian simulation therefore allows us to evaluate the impact of model misspecification. For both simulations, we generate 5000 samples and repeat this process 30 times with different simulation parameters to obtain average performance statistics for the different models.

\begin{table*}
    \centering
    \caption{Mean and standard deviation of the squared estimation error of population-level counterfactuals $p_{00},\dots,p_{11}$.}
    \vspace{2pt}
    \label{tab:se_pop}
    \begin{tabular}{l*{2}{S[table-format=1.2(1.2)e1]}}
        \toprule
        Model               & {Dirichlet simulation} & {Gaussian simulation} \\
        \midrule
        Independence & 1.79(5.62)e-3 & 1.25(0.32)e-2 \\
        Midpoint     & 1.16(3.63)e-3 & 1.91(0.55)e-2 \\
        BB           & 0.87(1.76)e-5 & 0.36(1.24)e-5 \\
        NBB          & 0.88(1.75)e-5 & 0.35(1.24)e-5 \\
        GBB          & 0.86(1.72)e-5 & 0.78(3.17)e-6 \\
        NGBB         & 0.88(1.75)e-5 & 1.14(3.32)e-6 \\
        \bottomrule
    \end{tabular}
\end{table*}

\begin{table*}
    \centering
    \caption{Mean and standard deviation of the squared estimation error of individual-level counterfactuals $p_{00}(x),\dots,p_{11}(x)$.}
    \vspace{2pt}
    \label{tab:se_cond}
    \begin{tabular}{l*{2}{S[table-format=1.2(1.2)e1]}}
        \toprule
        Model               & {Dirichlet simulation} & {Gaussian simulation} \\
        \midrule
        Independence & 0.33(1.12)e-2  & 1.46(0.56)e-2 \\
        Midpoint     & 0.23(0.85)e-2  & 2.23(0.71)e-2 \\
        BB           & 0.18(3.11)e-2  & 0.67(1.56)e-4 \\
        NBB          & {--}               & 0.72(1.45)e-4 \\
        GBB          & 0.14(2.78)e-2  & 0.69(1.91)e-4 \\
        NGBB         & {--}               & 0.71(1.38)e-4 \\
        \bottomrule
    \end{tabular}
\end{table*}

Table~\ref{tab:se_pop} reports the squared error between the true and estimated population-level counterfactuals for our four estimators, along with two baselines (independence and midpoint estimators) introduced in~\cite{verhelst2023partial}. Our proposed approach performs better than these two baselines, with an error lower by two orders of magnitude. The generalized beta approaches, GBB and NGBB, provide a slight increase in performance, which is especially noticeable on the Gaussian simulation.

The results for the estimation of individual-level counterfactuals are shown in Table~\ref{tab:se_cond}. We are not able to report their results for individual-level counterfactual estimation on the Dirichlet simulation of NBB and NGBB due to long inference time. We see that the baselines perform the worst, while GBB shows the best performance.

While the empirical analysis in this paper is restricted to simulated data, the same methods have already been applied successfully to real churn prevention campaign data in the telecom sector; see~\cite[Sec.~6]{verhelst2024causal} for a detailed evaluation. Together with the performance gains observed on our simulations, this prior evidence indicates that the proposed improvements are promising for practical applications. Inference of counterfactual probabilities exposes causal patterns and untapped intervention opportunities that are otherwise hidden, such as the maximum profits that a campaign could have generated under perfect targeting, or the number of \emph{do-not-disturb} customers~\cite[Sec.~6.6]{verhelst2024causal}.


\begin{footnotesize}
\bibliographystyle{unsrt}
\bibliography{references}
\end{footnotesize}

\end{document}